\documentclass{article}
\usepackage{graphicx} 
\usepackage{amsmath,amssymb,amsthm}
\usepackage{hyperref}
\usepackage{algorithm}
\usepackage{algpseudocode}
\title{A 
Point Process Model for Optimizing Repeated Personalized Action Delivery to Users}
\author{Alexander Merkov and David Rohde}

\usepackage{xcolor}
\newcommand{\Merx}[1]{\noindent\textcolor{brown}{#1}\marginpar{\textcolor{brown}{A}}}

\newcommand{\Dave}[1]{\noindent\textcolor{red}{#1}\marginpar{\textcolor{red}{D}}}

\newcommand{\MerxOff}[1]{}

\newcommand{\changedByMerxOK}[2]{{#2}}
\newcommand{\changedByMerxOff}[2]{{#1}}
\newcommand{\suggestedByMerx}[2]{\textcolor{magenta}{#1}{\textcolor{blue}{#2}}}
\newcommand{\suggestedByMerxOK}[2]{{#2}}
\newcommand{\suggestedByMerxOff}[2]{{#1}}
\newcommand{\commentByMerx}[2]{\textcolor{red}{#1}\footnote{\Merx{#2}}}

\newcommand{\commentByMerxOff}[2]{{#1}}
\long\def\deletedByMerx#1{\Merx{\\\\Some stuff has been deleted here, see the source.\\\\}}
\long\def\deletedByMerxOK#1{}

\newcommand{\Hs}{\mathcal{H}}   
\newcommand{\Ds}{\mathcal{D}}   
\newcommand{\Ue}{\mathcal{U}}   

\newcommand{\iu}{u}   
\newcommand{\ea}{f}   
\newcommand{\ee}{{\bar e}}   

\newcommand{\RNN}{\mathcal{R}}   

\newcommand{\One}{\mathbf{1}}   
\newcommand{\Ind}[1]{\One_{\left[#1\right]}}   

\long\def\comment#1{}
\newcommand{\Df}[1]{{\ifvmode\else\/\fi \em #1\/\index{#1}}}

\newcommand{\RR}{\mathbb{R}}	

\newcommand{\Exp}{\mathbf{E}}	

\newenvironment{example}[0]{{\ifvmode\else\\\fi\bf Example.\ }}{{\em \quad \qed}}

\begin{document}

\maketitle

\section{Introduction}\label{sectIntro}

This paper provides a formalism for an important class of causal inference problems and then goes on to propose solutions to some interesting special cases.  The problem of interest is that we have observations of $I$ users interacting with some system, e.g. people interacting with an advertising system.  The causal aspect of the problem occurs because the system will have an opportunity to intervene on each users possibly multiple times.  Additionally the system is able to personalize the actions being delivered to the users, there will be an initial observation of how the users behaves before any action has been delivered, in addition between each system actions
there will be additional observations that may be used to further personalize the delivery of future actions.

In section \ref{sectGeneral} we describe a very general idea of the proposed approach, too general to be implemented without further specification. In section \ref{sectTemporal} we show one of possible ways to specify the data accurately enough to give to this idea an exact formulation. In section \ref{sectNeural} we show one of possible classical technologies applicable to implement this approach in simple cases.

\section{General Idea}\label{sectGeneral}

Let $\iu$ identify a user, and let \suggestedByMerxOff{$\mathcal{H}_0^\iu$ be the information}{
$\Hs_0^\iu=(e^\iu_1,\dots,e^\iu_{N^\iu_0})$ be the time-ordered sequence of events $e^\iu_j$
}\ 
observed on user $\iu$ at the time the system has its first opportunity to act%
.  
The action delivered 
\suggestedByMerxOff{}{in response\footnote{It is possible to consider processes with actions arising independently of the events. It would be another problem. Here we consider the setting where an action cannot happen without certain event preceding. } }
is then denoted $a_1^\iu$.  After this action is delivered some more behavior of the user is observed, the information observed after the first action is delivered, and before the second request to act is denoted 
\suggestedByMerxOff{$\mathcal{H}_1^\iu$}{$\Hs_1^\iu=(e^\iu_{N^\iu_0+1},\dots,e^\iu_{N^\iu_1})$}, this is then followed by the second action $a_2^\iu$, and this sequence continues until \suggestedByMerxOff{some criterion is met}{the observation ends} and we have \suggestedByMerxOff{made $S^\iu+1$ observations, and delivered $S^\iu$}{observed $S^\iu+1$ subsequences of the sequence of $N^\iu_{S^\iu}$ events and $S^\iu$ delivered} actions\suggestedByMerxOff{}{\ which are events also}.  \changedByMerxOK{An example of such a criterion would be that the difference in time between 
\commentByMerx{the last event}{What is an event? You have rejected my attempt to define it in this section.
\suggestedByMerx{}{I suggest to admit that this section describes a preliminary large-scale picture missing some essential details and do not try to mention these details here.}
} \suggestedByMerxOff{in $\mathcal{H}_{S^I}^\iu$}{$e^\iu_{N^\iu_S}$} 
and
the first event \suggestedByMerxOff{in $\mathcal{H}_0^\iu$}{$e^\iu_1$} 
exceeds a pre-fixed maximum value $t_{\max}$.  
}{
}
Imagine that we have a set of logs 
$$\Ds = \{ \Hs^\iu_{0:S^\iu},a^\iu_{1:S^\iu} \}_{\iu=1}^I
= \left(\{ \Hs^\iu_{0:S^\iu}\}_{\iu=1}^I, \{ a^\iu_{1:S^\iu} \}_{\iu=1}^I \right) =:(H(\Ds),A(\Ds))
\footnote{By $X_{i:j}$ we denote the sequence $X_i,X_{i+1},\dots,X_{j}$, possibly empty if $j<i$.}
$$
\suggestedByMerxOff{}{$=\{ e^\iu_{1:N^\iu_{S^\iu}},a^\iu_{1:S^\iu} \}_{\iu=1}^I$}%
produced by $I$ independent stationary random processes with common probability distributions $P(\Hs_k|\Hs_{0:k-1},a_{1:k-1};\theta)$ (i.e. the probability of the observation\footnote{We have defined neither the space of observations, nor a probabilistic measure on it yet. It will be done in the next sections.} 
given its pre-history and the parameter)
\suggestedByMerxOff{}{with distribution $P(e^\iu_{k}|e^\iu_{1:k-1},a^\iu_{1:s(k)};\theta)$, where $s^\iu(k)=\min\{s:N^\iu_s\ge k\}$ is the number of the history segment $\Hs^\iu_s$ containing event $e^\iu_k$, within some parametric family of distributions (conditioned on the sequences of any length!)}
parameterized by $\theta$ of some parameter space $\Theta$. Then we can write the likelihood of the observations in this dataset given the actions there {and} this parameter:


\suggestedByMerxOff{
\commentByMerxOff{
\begin{align} 
& P(H(\Ds)|A(\Ds), \theta) = \prod_i^I P(\mathcal{H}_{S^\iu}^\iu|\mathcal{H}_{0:S^\iu-1}^\iu,a_{1:S^\iu}^\iu,\theta) \times ... \times  P(\mathcal{H}_1^\iu|\mathcal{H}_0^\iu,a_1^\iu,\theta) P(\mathcal{H}_0^\iu|\theta) 
\label{likelihood}
\end{align}
}{This equation is wrong in presence of the criterion $t_{\max}$ a few lines above. See also equation (\ref{eqPEventSeqBounded}) \suggestedByMerx{}{If we throw away time-dependent details from the above, this will come back true.}
}
}{
\begin{align} 
P(H(\Ds)|A(\Ds), \theta) &
                        = \prod_{\iu=1}^I 
                            \prod_{s=0}^{S^\iu} 
                            P(\Hs^\iu_s|\Hs^\iu_{1:s-1},a^\iu_{1:s},\theta)
\nonumber                            
\\ &                            
                        = \prod_{\iu=1}^I 
                            \prod_{s=0}^{S^\iu} 
                                \prod_{k=N^\iu_{s-1}+1}^{N^\iu_s}
                            P(e^\iu_k|e^\iu_{1:k-1},a^\iu_{1:s},\theta)
\nonumber                            
\\ &                            
                        = \prod_{\iu=1}^I 
                                \prod_{k=1}^{N^\iu_{S^\iu}}
                            P(e^\iu_k|e^\iu_{1:k-1},a^\iu_{1:s^\iu(k)},\theta)
\label{likelihood}
\end{align}
where the same parameter $\theta$ parameterizes a series of distributions on the sequences of events of any length conditioned on the sequences of events of any other length and sequences of actions of some third length.
}
\noindent
\changedByMerxOK{}{where the same rather abstract parameter $\theta$ parameterizes a series of distributions on the uncertain space of ``observations'' conditioned on sequences of any length of ``observations'' separated with ``actions''.}
Similarly%
\changedByMerxOK{}{, \commentByMerxOff{given some prior distribution $P(\theta)$}{A sweet dream\dots Can you imagine the space $\Theta$ of the parameters $\theta$?},}
the posterior can be written:

\begin{align}\label{eqPosterior}
P(\theta|\mathcal{D}) \propto P(\theta) P(H(\Ds)|A(\Ds), \theta)
\end{align}
\changedByMerxOK{}{up to proportionality.}


This model makes few assumptions other than stationarity%
\changedByMerxOK{
i.e. \commentByMerx{conditional on $\theta$ and $a_{1:S}$ the draws of $\mathcal{H}^\iu_{0:S^\iu}$ are independent}{This ``independence'' cannot have the standard sense that the joint probability of the two random values together is equal to the product of their probabilities. Then, what does it mean?
\suggestedByMerx{}{This can be formulated correctly in terms of policy introduced a dozen lines below.}
},
which could be relaxed by 
\commentByMerx{indexing the parameter $\theta$ by time}{
This has no definite sense (yet):
neither history segments $\Hs^\iu_s$, nor even individual events $e^\iu_k$ are bound to time (the events will get timestamps below);
such a time-dependent model cannot be learnt without rather heavy restrictions on its  time dependency.
\suggestedByMerx{}{I suggest to throw it away.}
}
}{}, and arrow of time assumptions, i.e. actions only impact \changedByMerxOK{events}{observations} after they were delivered, and the assumption that actions delivered to a users do not affect another users (an assumption sometimes called SUTVA \cite{rubin1986comment}).  This formulation is useful for determining how strongly past data supports or contradicts different values of the parameters, but it is inherently tied to the actions actually delivered.  Usually in practice we are interested in the strategy, algorithm or to deliver (personalized) actions.  We will adopt the convention of calling this algorithm a \emph{policy} and will write it as a probability distribution i.e. $\pi_\xi(a_s|a_{1:s-1},\Hs_{0:s-1})$ which denotes the probability of the $s$-th action personalized on the basis of $s-1$ previous actions \changedByMerxOK{}{and observations} \changedByMerxOK{and}{within a family of distributions} parameterized by some parameter $\xi$\changedByMerxOK{$\in$}{
\ within some parameter space\ 
}$\Xi$.  It should be noted that the policy is set by \changedByMerxOK{\commentByMerx{us}{Again: who are \emph{we} and who sets the policy?
\suggestedByMerx{}{I can change it if you do not mind.}}
}{the system} and is not something that \changedByMerxOK{\commentByMerx{we are uncertain about}{Here ``we'' are certainly ``the author and the readers'', it's OK, and knowing the policy and the past we might be certain, say, that certain action $\alpha_1$ will happen with probability $1/3$, and another action $\alpha_2$ --- with probability $2/3$.}, the probabilistic notation 
in most setting is just a convenience providing a relaxation when usually 
\commentByMerx{a deterministic decision rule is optimal}{but absolutely unsuitable for learning with Algorithm \ref{alg:cap} relying upon $\nabla_\xi\log\pi_\xi(\dots)$}}{should be predicted}.
The predictive distribution  for a policy $\pi_xi$ with parameters $\xi$ now becomes:




\[
 P(\Hs_{0:S}, a_{1:S} |\xi, \Ds)  = \left( \prod_{s=1}^S P(\mathcal{H}_s|\mathcal{H}_{0:s-1},a_{1:s},\mathcal{D})  \pi_\xi(a_s|
 a_{1:s-1},
 \mathcal{H}_{0:s-1}) \right) P(\mathcal{H}_0|\Ds)
\]

\noindent
where
\changedByMerxOK{}{$S+1$ is the number of observations, $S$ is the number of opportunities to act,}
\[
P(\mathcal{H}_s|\mathcal{H}_{0:s-1},a_{1:s},\mathcal{D}) = \int P(\mathcal{H}_s|\mathcal{H}_{0:s-1},a_{1:s},\theta) P(\theta|\Ds) d\theta.
\]

\noindent
and
\[
P(\mathcal{H}_0|\Ds) = \int P(\mathcal{H}_0|\theta) P(\theta|\Ds) d\theta
\]
\changedByMerxOK{}{are known up to proportionality.}

\noindent
\changedByMerxOK{
\commentByMerxOff{The optimization of the policy parameterized by $\xi$ requires that we specify a utility function, for an interesting class of problems it is sufficient to specify a utility on a single users $\iu$ by specifying a utility $U(\Hs_{0:S^\iu}^\iu,a_{1:S^\iu}^\iu) $}{This is true but can be understood only by by those acquainted with this terminology. \Dave{To explain this it could be a good idea to add an Algorithm 5 (probably in the appendix) where the utility is not on an individual user but on many users} 
\Merx{I'd rather give a definition of the utility right here, in the introduction. I am able to invent a utility function not additive in users, but then the set of users becomes a parameter of learning the policy. I doubt it is what you want.}
}}
To define optimization of the policy we need to define \emph{what} and \emph{under what condition} to optimize. We choose some real-valued \emph{utility function} $U$ defined on sequences of observations and actions $(\Hs_{0:S},a_{1:S})$ of any length $S$, the larger utility, the better. And we need to define some criterion how to start and stop the sequence to compute \emph{its utility} we are interested in. 

\begin{example}
For a toy criterion of fixed length $S$ of the sequence $(\Hs_{0:S},a_{1:S})$
we can specify 
the \emph{expected utility} under the predictive distribution 
\changedByMerxOK{}{$\theta$ of the policy $\pi_\xi$ for sequences of length $S$}
using:
\changedByMerxOK{}{
\begin{align}
\Ue(\pi_\xi;\theta,S) =  \Exp_{(\mathcal{H}_{0:S},a_{1:S}) \sim P(\mathcal{H}_{0:S},a_{1:S}|\xi,\theta)
                     }
                            U (\mathcal{H}_{0:S},a_{1:S})
\label{eqExpectedUtility}
\end{align}
or alternatively
}


\changedByMerxOK{
\commentByMerxOff{
\begin{align}
\Ue(\xi;\Ds) =  \Exp_{\mathcal{H}_{0:S},a_{1:S} \sim P(\mathcal{H}_{0:S},a_{1:S}|\xi,\Ds)
                     }
                            U (\mathcal{H}_{0:S},a_{1:S})
\label{eqExpectedUtility}
\end{align}
}{What is $S$ here? In fact, it is the key point why the $(\Hs,a)$ paradigm is not enough to compute something sensible.}
}{\begin{align}
\Ue(\xi;\Ds,S) =  \Exp_{(\mathcal{H}_{0:S},a_{1:S}) \sim P(\mathcal{H}_{0:S},a_{1:S}|\xi,\Ds)
                     }
                            U (\mathcal{H}_{0:S},a_{1:S})
\label{utility}
\end{align}
}
\end{example}   

\changedByMerxOff{}{Similarly the expected utility can be defined for other criteria, but it requires more elaborated notation, say, that we will introduce in section \ref{sectTemporal} below.}

\noindent
One way to maximize 
\changedByMerxOK{is the use of}{the expected utility is inspired by}
 the Reinforce algorithm \cite{williams1992simple} 
\changedByMerxOK{in order to access noisy derivatives}{namely, stochastic gradient maximization} of $\mathcal{U}(\xi;\Ds,C)$ with respect to $\xi$ for some end-of-sequence criterion $C$%
\changedByMerxOK{.}{\ using an approximation of the expected utility as the mean utility on the data sampled from the distribution learnt from available dataset $\Ds$.}  
\changedByMerxOK{}{
See rather informal algorithm \ref{alg:cap}
.
}
\begin{algorithm}[H]
\caption{\changedByMerxOK{Reinforce Algorithm}{stochastic gradient maximization of the expected utility\\
\# Convergence and ``End of Session'' conditions should be specified.\\
\# Gradient step parameter $\lambda$ should be described. 
}}\label{alg:cap}
\begin{algorithmic}
\State \changedByMerxOK{}{initialize $\xi$ \qquad \qquad\# to be specified\changedByMerxOff{}{\footnotemark}
}
\While{Not Converged}
    \State \changedByMerxOK{$\mathcal{H}_0 \sim P(\mathcal{H}_0|\Ds)$}{
        $\theta \sim P(\theta|\Ds)$  \qquad\#\ sampling from the distribution known only up to proportionality requires some explanation
        \State $\Hs_0 \sim P(\Hs_0|\theta)$
    }
    \State $s \gets 1$    
        \While{True}   
            \State $a_s \sim \pi_\xi(a_s|
            \changedByMerxOK{}{a_{1:s-1},}
            \mathcal{H}_{0:s-1})$  
                \changedByMerxOff{}{\qquad\#\ $\pi$ and $\xi$ will be declared soon} 
            \State $\mathcal{H}_s \sim P(\mathcal{H}_{s}|\mathcal{H}_{0:s-1},a_{1:s},\changedByMerxOK{\Ds}{\theta})$ 
            \If{End of Session Condition Met\changedByMerxOK{ in $\mathcal{H}_s$}{}}
                \State \changedByMerxOK{}{$S=s$;} break
            \EndIf
            \State $s \gets s + 1$    
        \EndWhile
        \State $\xi \gets \xi + \lambda$ 
        $U(\mathcal{H}_{0:S},a_{1:S})  \left(         \sum_{s=1}^S  \nabla_\xi \log  \pi_\xi(a_s|\mathcal{H}_{0:s-1},a_{1:s-1}) \right)$
        \State \changedByMerxOK{}{\qquad\#\ } the gradient step size $\lambda$\changedByMerxOK{}{$>0$ should be defined somehow}
    \changedByMerxOff{}{\State \#\ In contrast to true reinforcement learning, this simulation should ever stop
    and yield the learnt stochastic policy $\pi_\xi$ }
\EndWhile
\State \changedByMerxOK{}{return $\xi$}
\end{algorithmic}
\label{reinf}
\end{algorithm}
\changedByMerxOff{}{
\footnotetext{Will the parameter of the uniform distribution on the space of actions (which has neither name nor description) do? Does such a distribution belong to the family of the available policies, which also has neither name nor description?}
}


This formulation is simply Bayesian decision theory \cite{prevision} applied to a specific causal problem involving multiple opportunities to personalize treatment to multiple users under stationarity and SUTVA assumptions.  It could also be viewed as a Bayesian application of reinforcement learning.
In other respects we will also focus on a problem that also falls a little outside the traditional remit of reinforcement learning where both information we learn about the users and opportunities to act upon the users arise asynchronously.    

We will also assume that there is no unobserved confounding.  While this might seem like a strong assumption it is often absolutely satisfied, as it only requires that the actions are delivered using 
some policy $\pi_\xi(a_s|\Hs_{0:s-1})$, \changedByMerxOK{and not $\pi_\xi(a_s|\Hs_{0:s-1},u)$, where $u$ is some information used to deliver the action  but}{independent of anything} not available in the logs within $\mathcal{D}$.  Obviously by construction online interactive systems will not suffer from unobserved confounding\footnote{Poor practices such as having  hierarchies of models accessing different features can and do cause such systems to be susceptible to unobserved confounding, see \cite{rohde2024position} for more discussion.}.  We disagree with the impression given in \cite{wang2020causal,luo2024survey,spotify} that unobserved confounding is an unavoidable problem in interactive systems such as recommender systems, or computational advertising, and there is no need to use statistical methods to deal with confounding.

Furthermore, there is no use of any formalized \emph{causal inference} such as the do-calculus \cite{pearl1995causal}.  This is because \emph{causal inference is just inference} \cite{lattimore2019replacing,rohde2022causal}.  Similarly there is no need to use the propensity score \cite{rosenbaum1983central} either as a balancing score to remove confounding \cite{rosenbaum1983central} (for reasons already explained) nor as a means to compute the expected utility of the policy using the (non-Bayesian) Horvitz-Thompson estimator \cite{JMLR:v14:bottou13a}.  From a Bayesian perspective the propensity score should not be used, and indeed violates the likelihood principle \cite{berger1988likelihood}, although it's noteworth that this estimator is at the center of one of the most intense debates in statistics in recent times \cite{robins1997toward,sims2006example,robins2012robins}.

The model we have developed so far is extremely generic and no attention has been given to what type of information is observed in $\mathcal{H}_s^\iu$, and what type of action is denoted in $a_s^\iu$.  The rest of this document will develop models for cases inspired by interactive systems such as real time bidding and recommender systems, inspired in part by \cite{lewis2022incrementality}.  An important aspect of these problems is that \changedByMerxOK{all events are time-stamped i.e. both observations and opportunities to act occur asynchronously}{the observations are sequences of time-stamped events and some of the events cause actions}.  In order to handle this situation we will adopt the framework of neural temporal point processes \cite{shchur2021neural}, and we will discuss suitable models that simultaneously satisfy a) being a good model for a real system; b) have a tractable likelihood; and c) can be efficiently simulated from (enabling policy optimization). 

\comment{
Section 2.  Specific example of the above
The advertising engine is called upon to act upon a user when a request event occurs, which occurs at a random time.
A packet of events ends in one of two ways:
1) a request event
2) the end of session criterion occurs (e.g. discovery event has passed by t_max)
The first H_0 starts with discover user event.  The user becomes eligible for advertising, we discover the user exists.
We did not know the user existed before the event, and know nothing about their history.  An event occurs with a cookie that we have never seen before.
H_0 starts with a discovery event
H_s with s>0 can:
1) be empty (if time>time(discovery)+t_max
2) start with a click event, where the click is on a_{s-1} if time hasn't expired
3) start with any other event if time hasn't expired
4) can have arbitrary number of events, but has at most one request event which can only ever be the last event.
H_s conditional on all past history H_0:s-1 and actions a_s-1 is a marked temporal point process with a random finishing time. [controversial statement - to discuss!]

Damn! All these wishes require the notions of events, timestamps, etc., so such a section can be written only AFTER the current section 2 (Temporal Point Processes...). But then this H_s-notation will be hardly relevant. 
}

\section{Temporal Point Processes for Repeated Action Personalization}\label{sectTemporal}

We haven't yet specified what information is contained in $\mathcal{H}_s^\iu$, we will remedy this situation now, focusing on our problem class where both information and opportunities to act  
\changedByMerxOK{arrive asynchronously.}{are synchronized in a specific way. Our primary purpose is to formalize the timing and causation of what happens.
}

\changedByMerxOK{}{Observations $\mathcal{H}_s^\iu$ are time-ordered sequences of events $e^\iu_k$. Observation on the user $\iu$ begins at time $t^\iu_0$, either pre-defined or when a special event $e^\iu_0$ `the user appeared' happens, and goes on until time $t^\iu_0+t_{\max}$.
What happens to user $\iu$ after $t^\iu_0$ is a finite sequence of $B^\iu$ events $e^\iu_1,\dots,e^\iu_{B^\iu}$ of different types initiated by the user and actions made by the ``decision maker''\footnote{E.g. an advertising system.}. Each action is necessarily caused by an event of certain type and happens immediately after it\footnote{E.g. the user opens a web page of certain publisher and the advertiser decides to display (and displays!) some ads on its margins.}. So the actions will be considered as additional features of events of certain type(s) say, `request for action' or `request to display an ad'. For events of the types which cannot cause actions these additional features get some fixed value, say, $0$.
}

\changedByMerxOK{
The first piece of information  $\mathcal{H}_0^\iu$ contains $T^\iu_0$ events, each of these events contain both a timestamp and additional information (i.e. a `mark' $m$).  The `mark' contains an event type (denoted $v$) and (potentially) a high dimensional \changedByMerxOK{}{feature} vector $x$,  
\suggestedByMerxOK{}{so}
$m=(v,x)$%
.}{
Each event $e$ is featured with its timestamp $t$ (real), type $v$ (categorical), feature vector $x$ whose structure depends on the type: $e=(t,v,x)$, and it can be augmented with additional feature vector of the caused action: $\ee=(t,v,x,\ea)$. For terminological compatibility with marked temporal point processes (MTPP) the pair $m=(v,x)$ of the type and the features of and event will be called its mark.
}
\changedByMerxOK{
The type $v$ is a categorical variable, 
for our current purposes we will, for now, only define one of the categories: specifically the `request' type.  opportunities for 
\commentByMerxOff{us (the decision maker)}{I insist that usage of this pronoun in any sense except ``the author and the readers'' is misleading and should be avoided.} 
 to act on a users (usually a person) 
 \commentByMerxOff{occur asynchronously, we call an event requesting an action a `request event'}{
 Do you mean that each action is not asynchronous indeed but rather initiated by an event of some special type? Yes, the action can be delayed. But what can happen to the users within this delay? only nothing? any events? any events but request events? only request events? How many actions can be caused by a request event? by three request events? What an asynchrony!
 }.  
}{
Type $v$ which can take values like 'click an ad', 'view an item', etc. featured with the ad- or item-dependent features, and a distinguished value `request for action' featured with the action happened.
}


\changedByMerxOK{
We group together the first event(s) for users $\iu$ with 
\commentByMerxOff{the following notation}{
It is mostly the matter of taste, but the proposed notation looks very non-mnemonic and counter-intuitive: speaking about events characterized by their {\bf t}imes and {\bf t}ypes to denote their {\bf i}ndexes and {\bf n}umbers by $T^\cdot_\cdot$ and $B^\cdot_\cdot$ (see below). In my paper I used through numeration of the events and the translation will be rather painful and error-prone.
BTW, it seems, $t$ stands for time, but I am not sure. It has not been declared.
}:  $\mathcal{H}_0^\iu = \suggestedByMerxOK{}{e^\iu_{1:T^\iu_0} =} \{ (t,m)^\iu_{1:T_0^\iu} \} = \{ (t,v,x)^\iu_{1:T_0^\iu} \}$.  The final event type will be a $v_{T_0^\iu}={\rm `request'}$ type, and only the final event type is a request.  Similarly $\mathcal{H}_1^\iu$ contains $T_1^\iu$ events, with only the final event being of type request, and so on until $\mathcal{H}_{S^\iu-1}^\iu$  contain $T_{S^\iu-1}^\iu$ events, again ending on a request event.  Finally $\mathcal{H}_{S^\iu}^\iu$ will be terminated by some other criterion (perhaps a time window expiring).
}{
Let $B^\iu_0=0$, $B^\iu_1<\dots<B^\iu_{S^\iu}$ be the indexes of all events $e^\iu_\cdot$ of type `request for action', and $B^\iu_{S^\iu+1}=B^\iu$ be the index of the last event. They split the sequence $e^\iu_{1:B^\iu}$ of all such events into ${S^\iu}+1$ segments $
e^\iu_{({B^\iu_{s-1}+1}):B^\iu_{s}}
$, all of which, except possibly the last $({S^\iu}+1)$-th one, 
end with an event caused an action. These history segments consist of the ``information'' $\Hs^\iu_s$ mentioned in section \ref{sectIntro}. 
Segments of augmented events $
\ee^\iu_{({B^\iu_{s-1}+1}):B^\iu_{s}}
$
contain also the knowledge of the action $a^\iu_s$ followed, and they also explicitly contain the information on the timing. 
}


\changedByMerxOK{
In interactive systems problems that will be our focus it's useful to introduce another event type for `click', in this case for $\mathcal{H}_1^\iu$ to $\mathcal{H}_S^\iu$ 
if a `click' event occurs it will (usually) be the first within a grouping, but we will defer an in depth discussion of this to later sections.
\\
We will leave the definition of other event types, and the feature vector open for now as these definitions are highly dependent on the application.  Although it's worth noting that in recommender system applications there will be events where a users will reveal their interests or preferences, perhaps by engaging with an item or streaming a video etc.  If the action type is a recommendation, then this information will reveal what type of item(s) should be delivered as a recommendation.
\\
Finally the number of events for users $\iu$ is $B^\iu = \sum_{s=0}^{S^\iu} T^\iu_s$, and $S^\iu$ is also a random variable, similarly $B^\iu_s = \sum_{s'=0}^{s} T^\iu_{s'}$. 
}{} This notation lets us write 
\changedByMerxOK{
$\mathcal{H}^\iu_{0:s} = \{(t,m)^\iu_{1:B_s^\iu} \}$, and $\mathcal{H}^\iu_{0:S} = \{(t,m)^\iu_{1:B^\iu} \}$
}{
$\Hs^\iu_{0:S^\iu}=e^\iu_{1:B^\iu}
=(t,m)^\iu_{1:B^\iu}=(t,v,x)^\iu_{1:B^\iu}$
and \\
$(\Hs^\iu_{0:S^\iu},a^\iu_{1:S^\iu})=\ee^\iu_{1:B^\iu}=(t,m,\ea)^\iu_{1:B^\iu}=(t,v,x,\ea)^\iu_{1:B^\iu}  
$
, whichever detailization level we prefer%
}.



\suggestedByMerxOK{}{Now we can refine formula (\ref{likelihood}), algorithm \ref{alg:cap} and, for instance, turn the ``sampling $\Hs_s \sim P(\Hs_{s}|\Hs_{0:s-1},a_{1:s},\theta)$'' from a good wish in unspecified huge-dimensional space to somewhat modest-dimensional and implementable.}

\suggestedByMerxOK{}{\subsection{Marked Temporal Point Process: a probabilistic model}}

\suggestedByMerxOK{}{
Instead of rather abstract distributions $P(\Hs^\iu_s|\Hs^\iu_{1:s-1},a^\iu_{1:s},\theta)$ of section \ref{sectIntro} we can describe much simpler (but still rather complex) distributions of individual events $P(e_k|e_{1:k-1},\theta)$ conditioned on the preceding ones. For simplicity we suppose in this section that the space of features $x$ is finite so there are finitely many possible ``marks'' $m=(v,x)$. 
}

\suggestedByMerxOK{}{
To reflect the supposed independence of the distributions of the users and stationarity of the process, we remove the users from notation and provide either real or fictitious event $e_0$ of special type `start of observations' at time $t_0$. For each subsequent event its probability will be written not in terms of its absolute time $t_k$ but of its delay $\tau_k=t_k-t_{k-1}$ with respect to the previous event, though in conditions we keep absolute times\footnote{The distribution might depend on the day of week}. 
\begin{align}\label{eqPEvent}
    P(e|\ee_{0:k},\theta) = P((\tau,m)|(t,m,\ea)_{0:k},\theta)
\end{align}
Strictly speaking this formula (\ref{eqPEvent}) does not define a probability measure because the next event 
might never happen with positive probability. To fix it we add the `no-event' $e_\infty=(\tau_\infty,m_\infty)=(\infty,\infty)$ to the event space. But we will never include this `no-event' into an event sequence.
}

\suggestedByMerxOK{}{
Now the probability of the users's history seemingly can be written as the product of much simpler low-dimensional distributions
\begin{align}
    P(\Hs^\iu_{0:S^\iu}|a^\iu_{1:S^\iu},\theta)=P(e^\iu_{1:B^\iu}|a^\iu_{1:S^\iu},\theta)
    &=\prod_{k=1}^{B^\iu} P(e^\iu_k|\ee^\iu_{0:k-1},\theta) 
    \nonumber \\ &
    = \prod_{k=1}^{B^\iu} P((\tau,m)^\iu_k|(t,m,\ea)^\iu_{0:k-1},\theta)
    \label{eqPEventSeq}
\end{align}
}
\suggestedByMerxOK{}{
But this is not all the truth. Why had the sequence of events ended at $e^\iu_{B^\iu}$? Because the number $B^\iu$ was predefined, 
because `no-event' $e_\infty$ appeared,
or just because the observation time exceeded some predefined limit $t^\iu_0+t_{\max}$? In the latter case the probability depends also on $t_{\max}$:
\begin{align}\label{eqPEventSeqBounded}
    P(e^\iu_{1:B}|t_{\max},\theta)=&\prod_{k=1}^{B^\iu} P((\tau_k,m_k)^\iu|(t,m,\ea)^\iu_{0:k-1},\theta) 
\\ \nonumber &    
    \times \Ind{t_{B^\iu}\le t^\iu_0 + t_{\max}}
\\ \nonumber &    
    \times P(\tau>t^\iu_0 + t_{\max}-t^\iu_{B^\iu}|(t,m,\ea)^\iu_{0:{B^\iu}},\theta)
\end{align}
}

\suggestedByMerxOK{}{
We postpone considering still huge set of conditions $(t,m,\ea)_{0:k-1}$ and $\theta$ of the distribution $P((\tau_k,m_k)^\iu|(t,m,\ea)^\iu_{0:k-1},\theta)$ until section \ref{sectNeural} and first discuss the rather manageable space of distributions of pairs (time delay, mark) $(\tau,m)$.
}

\suggestedByMerxOK{}{
Provided the mark $m$ belongs to some finite set $\{1,\dots,M\}\cup\infty$, the probability of the event can be factorized as $P(e|\dots)=P((\tau,m)|\dots)=P(\tau|m,\dots)Q(m|\dots)$. The space of $(M+1)$-valued multinomial distributions is $M$-dimensional, so if we choose a $D$-dimensional family of distributions on $\RR_+$ for $\tau$ (for $m<\infty$), we get the $(D+1)M$-dimensional space $\Phi$ of distributions $P_\phi$, $\phi\in\Phi$ of pairs $(\tau,m)$. 
}

\suggestedByMerxOK{}{
For probability distributions from the space $\Phi$ all the factors of formula \ref{eqPEventSeqBounded} should be easily computable, they should better be easily differentiable in parameters to learn $\theta$ if needed, and they should allow easy sampling for future policy optimization (like in algorithm \ref{alg:cap}). All this assertion are trivial for multinomial distribution $Q_\phi(m|\dots)$ and might be non-trivial for time distributions $p_\phi(\tau|m,\dots)$. Usually such classic few-parametric families of distribution as log-normal (two-dimensional),  Erlang(a series of one-dimensional), gamma (two-dimensional) are tried for time distributions $p_\phi(\tau|m,\dots)$ of marked temporal point processes. In \cite{shchur2019intensity} it was shown that several-dozen-dimensional mixture of log-normal distribution gives better results. It was also conjectured in \cite{shchur2019intensity}, that the reason why the mixture is better is that all the base distributions have light tails, no heavier than exponential, while in real world the delay times are heavy-tailed.
}

\begin{example}
\suggestedByMerxOK{}{
We can suggest a non-standard three-dimensional family of heavy-tailed distributions on $\RR_+$ which allow both easy computation and sampling from. 
Each distribution of this family is unimodal and parameterized by its maximum point $\tau^*$, maximum value $p^*$ and the power speeds $\alpha$ and $\beta$ of decay approaching $0$ and $\infty$; of these $4$ parameters only $3$ are independent. Its density is
\begin{align} \label{eqPiecewisePowerDensity}
    p(\tau|\dots) = \left\{
                    \begin{array}{ll}
                         \left(\frac{\tau}{\tau^*}\right)^\alpha p^* & 
                         \mbox{when\  } 0\le\tau\le\tau^* \\
                         \left(\frac{\tau}{\tau^*}\right)^{-\beta} p^* & 
                         \mbox{when  } \tau\ge\tau^*
                    \end{array}
                \right.
\end{align}
where $\tau^*>0, \alpha>0$, $\beta>1, p^*>0$ and
\begin{align} \label{eqPiecwisePowerFullProb}
    \int_0^\infty p(\tau|\dots) d\tau = \left(\frac1{\alpha+1}+\frac1{\beta-1}\right) p^* \tau^*  =1
\end{align}
We can express, say $p^*$ via the other parameter using (\ref{eqPiecwisePowerFullProb}) and plug it into (\ref{eqPiecewisePowerDensity}); this gives us
\begin{align} \label{eqPiecewisePowerDensity3}
    p(\tau|\alpha,\beta,\tau^*) = \left\{
                    \begin{array}{ll}
                         \frac{(\alpha+1)(\beta-1)}{(\alpha+\beta)\tau^*} \left(\frac{\tau}{\tau^*}\right)^\alpha & 
                         \mbox{when\  } 0\le\tau\le\tau^* \\
                         \frac{(\alpha+1)(\beta-1)}{(\alpha+\beta)\tau^*} \left(\frac{\tau}{\tau^*}\right)^{-\beta} & 
                         \mbox{when  } \tau\ge\tau^*
                    \end{array}
                \right.
\end{align}
For such a density both the CDF 
\begin{align} \label{eqPiecewisePowerProbability3}
    P(\tau|\alpha,\beta,\tau^*) = \left\{
                    \begin{array}{ll}
                         \frac{\beta-1}{\alpha+\beta} \left(\frac{\tau}{\tau^*}\right)^{\alpha+1} & 
                         \mbox{when\  } 0\le\tau\le\tau^* \\
                         1 - \frac{\alpha+1}{\alpha+\beta} \left(\frac{\tau}{\tau^*}\right)^{1-\beta} & 
                         \mbox{when  } \tau\ge\tau^*
                    \end{array}
                \right.
\end{align}
and its inverse can also be expressed in elementary functions and computed fast, so computations of likelihoods and sampling are easy and maximization of the likelihood is possible\footnote{though with such inconveniences as non-smoothness and non-convexity}.
}
\end{example}

\suggestedByMerx{}{
}

\section{Neural Temporal Marked Point Processes}\label{sectNeural}

\suggestedByMerxOK{}{
After a family $\Phi$ of distributions $P_\phi$, $\phi\in\Phi$ on the space of events is chosen, to model the probability $P(e|\ee_{0:k},t_{\max},\theta)$ of an event given the previous events one can consider the parameter $\theta$ as a mapping from the space of finite time-ordered sequences $\ee_{0:k}$ of events augmented with consequent actions to $\Phi$. The set $\Theta$ of these mappings $\theta$ can be rather wide, and either an appropriate mapping $\theta^*\in\Theta$ or a posterior distribution on $\Theta$ can be learnt using some training datasets $\Ds$ and a likelihood-based method. Such an intention was declared in formulas (\ref{likelihood}) and (\ref{eqPosterior}). Now we try to implement it.
}

\suggestedByMerxOK{}{
As the space $\Theta$ we suggest recurrent neural networks (RNNs) in a broad sense: RNN, GRU, LSTM, Transformer, or whatever else. This approach is very popular (see survey \cite{shchur2021neural}) because of sequential nature of conditioning on $\ee_{0:k}$ in $P(e|\ee_{0:k},t_{\max},\theta)$. A neural network is defined by its particular structure (the number of neurons and layers, connections between them, activation functions), which should be chosen to define the space $\Theta$, and trainable weights which define a particular network $\theta\in\Theta$. We will not discuss the technically complicated but already classic problems of building and training of RNNs, and focus on the specifics of their application to the marked temporal point processes we are interested in.
}

\suggestedByMerxOK{}{
Let us remind the general structure and usage of recurrent neural networks.
Each RNN is a chain of identical subnetworks $\RNN(\theta)$ (briefly, $\RNN$), one for each element of the sequence to be processed, passing to each other some state vectors $\sigma_j\in\RR^d$ where $d$ is a structural parameter of the RNN, taking the previous augmented event $\ee_{j-1}$ and computing the parameter $\phi_j(\theta)$ (briefly $\phi_j$) of the distribution $P_{\phi_j}$ of the current event $e_j$
:
\begin{align}
\label{eqRNNstate0}      
    \sigma_0=&0
\\\label{eqRNNoutput}   
    \phi_j=&(\RNN.\phi)(\sigma_{j-1},\ee_{j-1}) \mbox{\qquad$j=1,2,\dots$}
\\\label{eqRNNstate}      
    \sigma_j=&(\RNN_.\sigma)(\sigma_{j-1},\ee_{j-1}) \mbox{\qquad$j=1,2,\dots$}
\end{align}
}


\suggestedByMerxOff{}{For augmented event $\ee_{j-1}=(\tau,m,f)_{j-1}$ the RNN $\RNN$ takes the 
feature vector $x_{j-1}$ into account, but the computed predictive distribution $P_{\phi_j}$ does not depend on $x_j$. If the feature vector is defined only for events of type `request for action' it corresponds to the sense of $x_{j-1}$ exactly. If for the other event types these vectors belong to small finite sets, such types can be just split into several subtypes with no feature vectors. In general case of multi-dimensional vector $x$ we have no good solution%
\changedByMerxOK{, and a partial workaround is presented in \label{pageSubtypes} section \ref{sectContextIndex}}{}. Forget about prediction of $x$ by now.
}

\suggestedByMerxOK{}{Algorithm \ref{algLikelihood} shows the current simplified approach to computation of the likelihood of the sequence of observed events within the specified time interval.
}
\suggestedByMerxOK{}{
If, say, 
$p(e|\phi)=p((\tau,m)|\phi)=p(\tau|m,\phi)Q(m|\phi)$ for some multinomial distributions $Q(\cdot|\phi)$ and $p(\cdot|\cdot)$ from the family (\ref{eqPiecewisePowerDensity3}), then the cumulative distribution $P(\tau|\phi)$ of all events is the mixture of the distributions $P(\tau|m,\phi)$ of shape (\ref{eqPiecewisePowerProbability3}) and can be computed easily.
}

\suggestedByMerxOK{}{
\begin{algorithm}[H]
\caption{Computation of the likelihood of the sequence of events within the specified observation time interval}\label{algLikelihood}
\begin{algorithmic}
\State  Input: function $p(e|\phi)$ computing the probability density of the event 
\State  Input: function $P(\tau|\phi)$ computing the cumulative probability of any event in the time interval $(0,\tau)$
\State  Input: trained RNN $\RNN(\theta)$ computing parameter $\phi$ of the distribution
\State  Input: the sequence $e$ of $B$ time-ordered augmented events, $e_k=(t,m,\ea)_k$
\State  Input: $t_0 \in \RR, t_{\max}\in\RR_+$ the start and the duration of the observation 
\State  Output: the conditional probability $P(e|t_0,t_{\max},\theta)$
\State
\If{$((e_1).t < t_0)\mbox{\ or\ }((e_B).t > t_0+t_{\max})$}     
    \State return(0)   \qquad \# the sequence does not fit into the observation interval
\EndIf
\State
\State  $k\gets0$;  $\sigma\gets0$ \qquad \qquad \# the current event number and the RNN state
\State  $e_0\gets(t_0,$'start'$,0)$  \ \qquad \#  the initial (zeroth) pseudo-event
\State  $r=1$   \qquad \qquad \qquad \qquad \# the probability to be returned
\While{$k\le B$}
    \State  $(\phi,\sigma)\gets \RNN(\theta)(\sigma,e_k)$   \quad \# compute the distribution and the next state
    \State  $k\gets k+1$   \qquad \qquad \qquad \# the next event number
    \If{$k\le B$}    \State  $r\gets r* p((t_k-t_{k-1},m_k)|\phi)$
    \Else            \State  $r\gets r* (1-P(t_0+t_{\max}-t_{k-1}|\phi))$    \qquad \# no next event observed
    \EndIf
\EndWhile
\State  return(r)
\end{algorithmic}
\end{algorithm}
}

\suggestedByMerxOK{}{Then the likelihood of the dataset $\Ds$ as a result of observation of the set of $I$ users within the time segments $[t^\iu_0,t^\iu_0+t_{\max}]$\footnote{
The time segment lengths $t^\iu_{\max}$ might be set to each user individually as well, but then they had to be given explicitly. They cannot be computed just as the maximal differences between the timestamps of events of the user available in the log.
}  given the actions $A(\Ds)$ happened
respectively is:
\begin{align}\label{eqLikelyhoodBounded}
P(\Ds|A(\Ds),t_{\max},\theta)
=
\prod_{\iu=1}^{I} & \left(
    \prod_{k=1}^{B^\iu} P_{\phi^\iu_k(\theta)}\left((\tau,v)^\iu_k\right) 
    \Ind{t^\iu_{B^\iu}\le t^\iu_0+t_{\max}}
    P_{\phi^\iu_{B^\iu+1}(\theta)}\left(\tau>t_{\max}-t^\iu_{B^\iu}\right)
\right)
\end{align}
(cf. motivational formula (\ref{likelihood}) and its refinement (\ref{eqPEventSeqBounded}); the details of computation of each of the $I$ top level factors are presented in algorithm \ref{algLikelihood}).
}

\suggestedByMerxOK{}{
Remind that $\theta$ is 
the vector of weights of a recurrent neural network which has independent on $\theta$ and already fixed structure, so either $\theta$ can be learnt\footnote{
Very roughly speaking, learning $\theta$, or equivalently, training the neural network $\RNN(\theta)$ given some dataset $\Ds$ is based upon maximization of the likelihood (\ref{eqLikelyhoodBounded}) or something like that. We are not ready to discuss training of RNNs of unspecified structure in details here. See, e.g. a textbook \cite{salem2022recurrent} 
} or the posterior on the set of these RNNs $\RNN(\theta)$ can be defined using $P(\Ds|\theta)$ as in formula (\ref{eqPosterior}). Finally, since all the distributions $P_\phi$ belong to a simple distribution class like (\ref{eqPiecewisePowerDensity3})\footnote{
More exactly, to class (\ref{eqPiecewisePowerDensity3}) belong type-conditioned time distributions $p(\tau|m,\dots)$ while the overall time distribution $p(\tau|\dots)$ is a mixture of such distributions.
}, sampling each next event $(\tau,m)$ is easy, see algorithm \ref{algSimpleSample}. 
}

\suggestedByMerxOK{}{
\begin{algorithm}[H]
\caption{Sample an event from a distribution $P((\tau,m))=p(\tau|m)Q(m)$ for multinomial $Q(\cdot)$ and $p(\cdot|\cdot)$ from the family (\ref{eqPiecewisePowerDensity3})--(\ref{eqPiecewisePowerProbability3}) }\label{algSimpleSample}
\begin{algorithmic}
\State  Input: $M$, $Q=(q_1,\dots,q_M)$,    \qquad $\sum_{m=1}^M q_m\le1$
\State  Input: $\alpha_m$, $\beta_m$, $\tau^*_m$, $m=1,\dots,M$ \qquad see (\ref{eqPiecewisePowerDensity3})
\State  Output: the sampled event $e=(\tau,m)$
    \State
    \State \#\#\#\#\#\#\#\# sample $m$ from multinomial $Q(m)$:
    \State $\eta \sim U([0,1])$ \qquad \qquad \# uniform sampling
    \State $m\gets 0$, $s\gets 0$;
    \While{ $s\le\eta$} 
    \If{$m=M$}
        \State  return{$(\infty,\infty)$}   \qquad \# no more events
    \EndIf
      \State $m\gets m+1$, $s\gets s+q_m$;
    \EndWhile   \qquad \# sampling in $O(\log M)$ time is also possible\footnotemark
    \State
    \State \#\#\#\#\#\#\#\# sample $\tau$ from $P(\tau|m)$, namely from (\ref{eqPiecewisePowerProbability3}):
    \State $\eta \sim U([0,1])$ \qquad \qquad \# uniform sampling
    \If{$\eta< \frac{\beta_m-1}{\alpha_m+\beta_m}$} \qquad \# compute the inverse $\left(P^{-1}(\tau|m)\right)(\eta)$
        \State  $\tau=\tau^*_m \left(\frac{\eta(\alpha_m+\beta_m)}{\beta_m-1}\right)^{\frac1{\alpha_m+1}}$
    \Else 
        \State  $\tau=\tau^*_m \left(\frac{(1-\eta)(\alpha_m+\beta_m)}{\alpha_m+1}\right)^{-\frac{1}{\beta_m-1}}$
    \EndIf
\State
\State return($\tau$,$m$)
\end{algorithmic}
\end{algorithm}
\footnotetext{but useless in this algorithm because is used only once for the given $Q$}
}

\suggestedByMerxOK{}{Now, after the RNN model $\RNN(\theta)$ and training its parameter $\theta$ are more or less described and sampling an event from it is shown explicitly, we can show how to sample the event sequences and learn the policy (as announced in algorithm \ref{alg:cap} rather informally). See algorithms \ref{algSeqSample} and \ref{algPolicyGradientOptimization}.
}

\suggestedByMerxOK{}{
\begin{algorithm}[H]
\caption{Sample a sequence of augmented events of the desired time duration 
}\label{algSeqSample}
\begin{algorithmic}
\State  Input: a RNN $\RNN$
\State  Input: function sampleNext($\phi$) (algorithm \ref{algSimpleSample})
\State  Input: $t_0 \in \RR, t_{\max}\in\RR_+$ the start and the duration of the sampling 
\State  Input: policy $\pi_\xi$ to be applied to events of type 'request for action'
\State  Output: the length $B$ and the sampled sequence $e_{1:B}=(t,m)_{1:B}$
\State
\State  $t\gets t_0$; $B\gets0$;  \qquad \# the current time and the length of the event sequence
\State  $\sigma\gets0$; $e_B\gets(t_0,$'start'$,0)$  \qquad \# the RNN state and the current event
\State
\While{$t<t_0+t_{\max}$}        
    \State  $(\phi,\sigma)\gets \RNN(\sigma,e_B)$   \qquad \# compute the distribution and the next state
    \State  $(\tau,m)\gets$ SampleNext$(\phi)$   \quad \# sample the next event from the distribution
    \State  $t\gets t+\tau$        \qquad \qquad \qquad \qquad \qquad \# the event absolute time
    \If{$t>t_0+t_{\max}$} break;   \qquad \# the time is over
    \EndIf
    \State $B\gets B+1$
    \State $e_B \gets (t,m,0)$           \qquad \# store the new event with no action
    \If{$m=$`request for action'}   
        \State  $a \sim \pi_\xi(e_{0:B})$   \qquad \# sample the requested action from the policy
        \State  $(e_B).\ea \gets a$             \qquad \# and store it as caused by this event
    \EndIf
\EndWhile
\State
\State return($B,e_{1:B}$)
\end{algorithmic}
\end{algorithm}
}

\suggestedByMerxOK{}{
\begin{algorithm}[H]
\caption{Stochastic gradient maximization of the policy utility}\label{algPolicyGradientOptimization}
\begin{algorithmic}
\State  Input: trained RNN $\RNN$
\State  Input: function sampleEvents($\RNN, t_0, t_{\max},\pi$) (algorithm \ref{algSeqSample})
\State  Input: $t_0 \in \RR, t_{\max}\in\RR_+$ the start and the duration of the sampling\footnotemark 
\State  Input: initial policy $\pi_{\xi_0}$ 
\State  Input: utility function $U$ defined on the event sequences $e_{1:B}$ 
\State  Input: gradient step parameter $\lambda>0$
\State  Input: a StopOptimization(\dots) criterion
\State  Output: the parameter $\xi$ of the optimized policy $\pi_{\xi}$
\State
\State  $\xi\gets\xi_0$
\State
\While{not StopOptimization(\dots)}        
    \State  $(B,e)\gets$ sampleEvents$(\RNN, t_0, t_{\max},\pi_\xi)$  
    \State  \qquad \qquad \qquad \qquad \qquad \ \# sample an event and action sequence $\ee_{1:B}$
    \State  $\xi\gets\xi+\lambda U(e)\left(\sum_{k=1}^B \Ind{m_k=\mbox{'request for action'}}\nabla_\xi\log\pi_\xi(\ea_k|e_{1:k-1})\right)$
    \State   \qquad \qquad \qquad \qquad \qquad \ \# a gradient step in the policy parameter space
\EndWhile
\State
\State return($\xi$)
\end{algorithmic}
\end{algorithm}
}
\footnotetext{In fact, they have no need to be fixed and might be changed on the run.}

\bibliographystyle{plain}
\bibliography{refs}

\end{document}